# Deep Learning for Grading Endometrial Cancer

Manu Goyal[*], PhD, Department of Biomedical Data Science, Dartmouth College, Hanover, NH, USA

Laura J. Tafe, MD, Department of Pathology and Laboratory Medicine, Dartmouth-Health, Lebanon, NH, USA

James X. Feng, Geisel School of Medicine, Dartmouth College, Hanover, NH, USA

Kristen E. Muller, DO, Department of Pathology and Laboratory Medicine, Dartmouth-Health, Lebanon, USA

Liesbeth Hondelink, MD, PhD, Department of Pathology, Leiden University Medical Center, Leiden, The Netherlands

Jessica L. Bentz, MD, Department of Pathology and Laboratory Medicine, Dartmouth-Health, Lebanon, NH, USA

Saeed Hassanpour, PhD, Departments of Biomedical Data Science, Computer Science, and Epidemiology, Dartmouth College, Hanover, NH, USA

**Short title:** Transformer for Endometrial Cancer

**Conflicts of interest:** The authors have no financial, professional, or personal conflicts of interest.

**Funding sources:** This research was supported in part by grants from the US National Library of Medicine (R01LM012837 & R01LM013833) and the US National Cancer Institute (R01CA249758).

**Number of text pages:** 16

**Number of Figures:** 3

**Number of tables:** 4

**\*Corresponding Author:** Manu Goyal, 1 Medical Center Drive, Williamson Translational Research Building, Lebanon, NH, 03756. Manu.Goyal@Dartmouth.edu, +1 (603) 678-9293

# Abstract

Endometrial cancer, the fourth most common cancer in females in the United States, with the lifetime risk for developing this disease is approximately 2.8% in women. Precise histologic evaluation and molecular classification of endometrial cancer is important for effective patient management and determining the best treatment modalities. This study introduces EndoNet, which uses convolutional neural networks for extracting histologic features and a vision transformer for aggregating these features and classifying slides based on their visual characteristics into high- and low- grade. The model was trained on 929 digitized hematoxylin and eosin-stained whole-slide images of endometrial cancer from hysterectomy cases at Dartmouth-Health. It classifies these slides into low-grade (Endometroid Grades 1 and 2) and high-grade (endometroid carcinoma FIGO grade 3, uterine serous carcinoma, carcinosarcoma) categories. EndoNet was evaluated on an internal test set of 110 patients and an external test set of 100 patients from the public TCGA database. The model achieved a weighted average F1-score of 0.91 (95% CI: 0.86–0.95) and an AUC of 0.95 (95% CI: 0.89–0.99) on the internal test, and 0.86 (95% CI: 0.80–0.94) for F1-score and 0.86 (95% CI: 0.75–0.93) for AUC on the external test. Pending further validation, EndoNet has the potential to support pathologists without the need of manual annotations in classifying the grades of gynecologic pathology tumors.

# Introduction

Endometrial cancer is the most prevalent cancer affecting the female reproductive tract in the United States [1]. The American Cancer Society estimates around 65,950 new cases of endometrial cancer will be diagnosed in 2023 in the US, and the disease is expected to result in approximately 12,550 deaths [1]. This malignancy, more common in postmenopausal women, exhibits a higher incidence rate in Black women, who also face a higher mortality rate [https://www.cancer.org/cancer/endometrial-cancer/about/key-statistics.html Last Accessed: 05 February 2024]. Globally, endometrial cancer ranks within the top 10 most common cancers [2].

According to the 2023 International Federation of Gynecology and Obstetrics (FIGO) staging system [3], Endometrial cancers are divided into type I (low-grade (grade 1 or 2) endometrioid), which typically have a good prognosis, and type II (grade 3 endometrioid, serous, clear cell, carcinosarcoma, undifferentiated/dedifferentiated) which have a relatively poor prognosis [4,5]. These tumors arise via different biological pathways with distinct molecular alterations [4,5]. The current standards for optimal management of endometrial cancer patients incorporate molecular classification in accordance with the guidelines set by the World Health Organization (WHO) [6], the European Society of Gynecological Oncology (ESGO) [7], and 2023 FIGO [3]. Molecular classification of endometrial cancer, as encouraged by WHO, ESGO and 2023 FIGO guidelines, offers a more precise prognosis and tailored treatment approach compared to traditional grading systems. However, its implementation is not yet widespread, particularly in less developed countries, due to challenges in resources and access to advanced diagnostic tools. Unlike the grading system, which assesses histopathological features like cellular atypia and tumor architecture, molecular classification involves analyzing specific genetic and molecular changes (POLEmut, MMRd, NSMP, p53abn) within cancer cells to guide treatment decisions.

The grading of endometrial cancer depends on the extent of cellular atypia and architectural complexity, which plays a crucial role in determining patient treatment. However, these grades demonstrate substantial intra- and inter-observer variability, leading to potential discrepancies in treatment [8]. Classification of endometrial cancer by histology alone has shown rates of interobserver disagreement ranging from 10% to 20% and up to 26% to 37% in high-grade tumors [8,9,10]. This variability can, in part, be explained by overlapping histologic features and morphologic ambiguity. Cancer staging, grading, and classification, which is based on histopathologic characterization, is important for patient treatment planning.

Automated image analysis using deep learning techniques can provide fast, accurate, and consistent results. These computational methodologies have been investigated in prior studies of endometrial cancer, including those focusing on hematoxylin and eosin-stained (H&E) histology slides, immunofluorescence images, and MRI scans [11-13]. Some previous work has observed a shift in endometrial cancer diagnosis, which is transitioning from a histological investigation to a more molecular based one and identified deep learning as a potentially powerful tool for holistic analysis of imaging and molecular data [14]. Machine learning-directed diagnostics could also aid in assigning clinical relevance to personalized histologic and molecular features within individual tumors [15].

Previously, a deep learning approach was applied by Kather et al. in the classification of multiple cancer types, including endometrial cancer. This study used a convolutional neural network (CNN) on digitized H&E tissue samples and demonstrated the utility of such models in histopathological analysis of cancer [16]. In another study, a multi-resolution CNN model named "Panoptes" showed promising results in predicting endometrial cancer subtypes and molecular classification. The model's performance on independent test sets was noteworthy, yielding an

AUROC of 0.97 for differentiating between endometrioid and serous subtypes and 0.93 for identifying high-risk CNV-H (copy number variant–high) molecular subtype [17]. The development of im4MEC, an interpretable deep learning model, was reported in another study. This model could predict molecular classes (e.g., *POLE*, *MMR* deficient, *TP53* mutated) of endometrial cancer based on Whole Slide Images (WSIs), offering improved prognoses and the identification of morpho-molecular correlates [18]. Furthermore, deep learning has also proven effective in categorizing cervical and endometrial cancer subtypes and determining the origin of adenocarcinomas [19]. These studies suggest that with increased clinical validation and integration, deep learning models could potentially facilitate enhanced discrimination of different cancer types to improve treatment planning [20-23].

A comprehensive analysis of key histologic features, including nuclear atypia, architectural features, variations in cytoplasmic qualities, and nucleoli visibility, plays an important role in histopathologic diagnosis of endometrial carcinoma. These features are important in differentiating the histologic types of endometrial carcinoma, a crucial step that informs diagnosis and treatment. The complexity and considerable heterogeneity in histopathologic features of endometrial cancer make its classification into distinct grades challenging but essential. Nevertheless, microscopic evaluation of histology slides is not only labor-intensive, but due to the histologic ambiguity of some tumors, it can also lead to variability in grading and classification among pathologists, potentially resulting in suboptimal clinical decision-making and treatment. Furthermore, smaller laboratories may not have access to ancillary testing, which could assist with more accurate distinction among tumor variants. This provides an opportunity for innovative solutions to enhance diagnostic accuracy and efficiency, facilitating personalized patient care. It is important to note that most preceding image analysis studies overlook the spatial context and relationships among the patches, which could limit comprehensive feature recognition, integration, and overall accuracy of the histologic classification [24, 25, 26].

To address these challenges in endometrial cancer classification, our study introduces EndoNet, a deep-learning model inspired by MaskHIT [27, 28]. EndoNet leverages the strengths of MaskHIT, which uses transformer outputs to reconstruct masked patches to learn local and global histologic features based on their positions and visual attributes. MaskHIT architecture forms the foundation for the histologic classification of endometrial cancer by improving visual recognition and differentiation between grades. This approach is focused on accurate histologic grade classification to assist clinicians in providing effective and personalized treatment plans.

## Materials and Methods

## Datasets

We used an internal dataset for training and testing EndoNet and further externally validated EndoNet on 100 randomly selected digitized slides from The Cancer Genome Atlas (TCGA) public database [29]. This study and the use of human subject data in this project were approved by the Dartmouth-Health Institutional Review Board (IRB) with a waiver of informed consent. The details of these datasets are included below.

**Internal Dataset.** This dataset contains 1143 WSI, corresponding to 549 patients, from the Department of Pathology and Laboratory Medicine at Dartmouth–Health, a tertiary academic care center in Lebanon, New Hampshire. These H&E-stained WSIs were randomly selected from routinely scanned hysterectomy specimen slides in the Department of Pathology and Laboratory

Medicine between 2016 and 2019. These slides were digitized using the Leica Aperio AT2 and CS2 scanners at 40× magnification (0.25 μm/pixel).

**The Cancer Genome Atlas (TCGA) dataset**. The TCGA dataset contains 560 digitized H&E tumor sample slides and corresponding molecular data. This dataset was collected from 373 patients, comprising 307 endometrioid and 66 serous or mixed histology cases, with acquisition approved by local institutional review boards [30]. For this study, we randomly selected 100 WSIs from 100 unique patients from TCGA for external validation in this study. These slides scanned at two distinct resolutions: 85 of the slides were scanned at a 40x magnification (0.25 μm/pixel), while the remaining 15 slides were scanned at a 20x magnification (0.5 μm/pixel).

**Data Annotation.** The histologic subtype of each WSI in the internal dataset was manually extracted from their associated pathology reports from the Department of Pathology and Laboratory Medicine at Dartmouth-Health. Based on this review, 214 slides were subsequently excluded from the study, which included slides with no malignancy, lymph nodes, and less commonly reported variants, such as clear cell carcinoma. Based on the consultation with our expert pathologist collaborators, because of the morphologic intratumorally heterogeneity that can be seen within one tumor, some slides may have more abundance of GR2 vs. GR3 areas of tumor and overall, those cases were classified as GR3 in this study. The distribution of these WSIs used in our Internal DH and TCGA dataset is summarized in Table 1. Ultimately, we utilized 929 slides, comprising 654 slides of endometroid grade 1 and 2, and 275 slides of a combination of endometroid FIGO grade 3, serous carcinoma, and carcinosarcoma. We downscaled WSIs from 40x to 10x magnification (i.e., 1 μm/pixel) using the python's openslide library function named Image.ANTIALIAS interpolation technique to optimize computational processing while preserving histologic features and details.

We also reviewed and labeled 100 randomly selected TCGA WSIs in the external test set. For these slides, the initial labels for endometrioid and serous carcinoma were determined using metadata from the TCGA database. For this study, we have only included 100 WSIs of 100 distinct patients out of 560 WSIs of 373 patients of TCGA cohort. Of note, TCGA metadata does not include the specific classes used in this study. Therefore, two expert gynecologic pathologists subsequently reviewed the slides and their labels, and detailed labels delineating histologic subtypes for Endometrioid FIGO grades 1, 2, and 3, along with serous carcinoma, were assigned by their corroborated opinion. In addition, any discrepancies were resolved through further review and discussion. Therefore, the time requirement from pathologists for the review and establishing the ground truth for these slides was the limiting factor on the number of included slides from TCGA in our external dataset. Based on this review, the external TCGA dataset contained 70 low-grade and 30 high-grade slides.

We further annotated a portion of slides in the internal training set to refine the features derived from CNN models for distinguishing endometrial cancer characteristics. For this purpose, three subspecialty gynecologic pathologists manually annotated randomly selected 187 WSIs in training by putting bounding boxes around tumors and classifying them by one of the labels listed in Table 1 using the Automated Slide Analysis Platform (ASAP) [https://github.com/geertlitjens/ASAP. Accessed 17 October 2022]. This bounding-box region of interest annotation was performed on the digitized WSI at the highest resolution (40x) and subsequently divided into smaller patches at 10x magnification.

**Dataset Partitioning for Model Development and Evaluation.** We partitioned the internal slides into subsets for model development and evaluation. A subset of 648 slides of 354 patients from the internal dataset (approximately 70% of the total internal dataset) served as the training set. We allocated 93 internal slides (roughly 10% of this dataset) to form a validation set and dedicated a collection of 188 internal slides from 110 patients (about 20% of this dataset) for the internal testing set as shown in Table 2. Careful attention was given to maintaining patient consistency within each set, ensuring that slides originating from the same patients were kept within the same subset to prevent data leakage. Moreover, we ensured the histologic composition mirrored the training/validation/testing split to uphold coherence and comparability across the different partitions. To further extend our evaluation, we used all the randomly selected slides from the TCGA database as the external test set. This external test set helped to evaluate the generalizability of our approach.

## Fully Supervised CNN and EndoNet Pipeline

For this study, we used two methodologies for the classification of low- and high-grade endometrial cancer. First methodology is based on the Fully Supervised CNN based approach which was modeled and tested on the previous works [24, 25, 26]. First, the patch-level classifier is trained on 224×224 patches of low- and high-grade endometrial cancer from the regions manually annotated by pathologists on the 187 WSIs. The details of the number of patches for each class are provided in Supplementary Table 1. The model is trained on selected patches from annotated areas with a 1/3 overlap, also using the color intensity normalization and on-the-fly data augmentation techniques such as random flips and color jittering to enhance model robustness. The training involves over 230,000 patches, with a balanced class distribution achieved through weighted random sampling. The ResNet-18 model is initialized using a normal distribution and is trained with cross-entropy loss over 100 epochs. It starts with an initial learning rate of 0.01, which is gradually decreased throughout the epochs. This model is then used to predict the unannotated patches to get the number of patches that belong to each class of the WSIs of the training, validation, and testing sets (same as those presented in Table 2). Finally, the logistic regression method is trained on the training set (by fine-tuning the different hyperparameters on the validation set) and providing the whole-slide inference on both the internal test set and the external TCGA set.

In our second methodology, we introduced EndoNet, inspired by previous work from our team on masked pre-training for histology images using a transformer (MaskHIT) to analyze endometrial cancer WSIs [27, 28]. This pipeline uses state-of-the-art techniques to capture fine and broad details within WSI and is trained to fill in missing or obscured parts of the image accurately. This functionality can aid with the representations of histomorphologic features and can differentiate between low and high-grade cases of endometrial cancer in this application. Moreover, this approach identifies the areas of interest within the image that most influence the classification. These influential regions can be visualized and help pathologists investigate which parts of the WSI were most significant in reaching a diagnosis, explaining the model outcome, and facilitating informed clinical decision-making.

This proposed methodology effectively combines low- and high-level features extracted from the images. Low-level features are obtained using a pre-trained CNN model on tissue patches, while high-level features are represented using a vision transformer model. This hybrid approach captures detailed local information and global relationships among patches. Figure 1 provides an overview of the EndoNet processing pipeline. In this study, two pre-trained ResNet18 models, trained on the ImageNet dataset and EC patches from a small subset of the training set (Fully Supervised CNN), were utilized to extract features from patches. Notably, these models were in a frozen state during the feature extraction phase. For pre-training and fine-tuning, the tissue segmentation operation is performed in the preprocessing step to exclude blank regions.

For this study, 25 regions from the WSI of endometrial cancer were selected and processed by fine-tuning the pre-trained transformer model, which was trained for cancer subtype classification of TCGA breast cancer in a prior study [28]. There was no masking applied to the patches in this fine-tuning process. The model-generated 'class tokens' for all regions were subsequently averaged through aggregation method to create a comprehensive representation of the entire slide. This averaged class token was then used to predict the classification of the slide, specifically whether it fell into the low-grade or high-grade category for endometrial cancer. The overview of training the EndoNet architecture is demonstrated in Figure 1.

**Evaluation Metrics and Statistical Analysis**

To assess the performance and effectiveness of our methodology, we evaluated our trained model on a held-out test set of 188 WSIs of 110 patients from our internal dataset and an additional 100 WSIs of 100 patients from the external TCGA dataset. The evaluation was based on two key metrics: the Area Under the Curve (AUC) and the F1 score. Furthermore, to F1 score statistical measure of these metrics' reliability, we calculated the 95% confidence intervals (CIs) using the bootstrap resampling technique with 10,000 iterations. The F1-score delivers a balanced evaluation of precision (the ratio of accurately positive results among all positive predictions) and recall (the ratio of accurately positive results among all genuine positive instances). High sensitivity and specificity are particularly important in the medical image analysis, as these metrics directly influence the patient outcomes and treatment plans. High sensitivity is crucial to ensure that most patients with the disease are correctly identified, whereas high specificity is necessary to minimize the number of healthy individuals incorrectly diagnosed with the disease. Conversely, the AUC score quantifies the model's capacity to differentiate between low-risk and high-risk endometrial cancer cases; a higher AUC value indicates superior performance. Collectively, these four metrics offer a comprehensive evaluation of EndoNet's efficacy in predicting endometrial cancer histologic classification.

## Results

**Classification of Low- and High-grade Endometrial Cancer**

In this study, we used the two methodologies i.e. the Fully Supervised CNN and the EndoNet for the classification of low- and high-grade endometrial cancers. We also experimented the EndoNet with two different pre-trained CNN models to find the optimal model for feature extraction. One model is pre-trained on the ImageNet dataset, while the other uses patches from manually annotated regions representing low-grade and high-grade regions in endometrial cancer WSIs, as was described above. The aim was to compare the effectiveness of these two different feature extractors in classifying endometrial cancer as low- or high-grade. Next, 80% of these patches were used to train the patch-level ResNet18 classifier, while the remaining 20% were used for validation. The most efficient instance of the pre-trained ResNet18 was chosen based on the epoch during which the model achieved the best AUC and F1 Score (0.93 and 0.91, respectively) on the patches from the validation set.

Table 3 summarizes the performance metrics and their corresponding 95% confidence intervals for classifying low- and high-grade patients in internal and external test sets of 110 and 100 patients respectively. The performance metrics are calculated on a per-patient basis by averaging the score for each patient. Additionally, an optimal threshold is used to maximize the difference between the true positive rate (sensitivity) and the false positive rate (1 – specificity). This optimal threshold is important for classification problems to distinguish between classes by maximizing both sensitivity and specificity.

The Fully Supervised CNN approach, utilizing the ResNet18 model, showed an F1 score of 0.78 and an AUC score of 0.82 on the internal test set of 110 patients, with sensitivity at 0.90 and specificity at 0.67. On the external TCGA test set of 100 patients, this approach achieved an F1 score of 0.86, an AUC of 0.79, a sensitivity of 0.57, and a specificity of 0.90. The EndoNet model, using the ResNet18 pre-trained on ImageNet as compared to Fully Supervised CNN approach, achieved an F1 score of 0.90 and an AUC score of 0.87 on the internal test set. It also achieved a high specificity of 0.91 and a moderate sensitivity of 0.72. On TCGA cohort, this model achieved an F1 score of 0.79, an AUC score of 0.83, a sensitivity of 0.87, a specificity of 0.69. On the other hand, when employing the ResNet18 model pre-trained on the annotated endometrial cancer patches in the proposed method, the F1 score increased to 0.91 with a corresponding increase in the AUC score to 0.95 on the internal test set. For sensitivity and specificity, the model achieved the score of 0.88 and 0.90 respectively. Finally, this model scored an F1-score of 0.86 and an AUC value of 0.86, alongside a sensitivity of 0.87 and a specificity of 0.80 on the external TCGA test set. The ROC curves of our proposed methods on both internal test set and external TCGA dataset are shown in Figure 2.

We further investigated performance of EndoNet in diagnosing different histologic types of endometrial cancer on both internal and external test sets (Table 4). For the internal testing set, while EndoNet demonstrated commendable accuracies of 0.94 for Endometrioid FIGO grades 1, 0.91 for grade 3 and 0.88 for carcinosarcoma respectively. However, there was a moderate decline in accuracy to 0.81 for Endometrioid FIGO grade 2 and to 0.85 for Serous Carcinoma. The TCGA dataset analysis confirmed EndoNet's reliable performance for Endometrioid FIGO grades 1 with an accuracy of 0.85 and for grade 3 with 0.88, alongside 0.83 for Serous Carcinoma. Yet, it again highlighted a noticeable performance decrease for grade 2 tumors to 0.68, which exhibited the lowest accuracy.

## Visualization

We randomly selected the ten cases from both internal and external test sets to verify the EndoNet's attention in the classification of low- and high-grade endometrial cancer. Figure 3 shows representative samples from internal and TCGA testing datasets, featuring the attention maps derived from the EndoNet model as it is tasked with binary classification of low-grade and high-grade endometrial cancer. These attention maps function as a visual representation of the model's focus while making its classification decisions and showed notable concordance within the tumor regions, as annotated independently by our expert gynecologic pathologists.

## Discussion and Future Directions

Endometrial cancer ranks as the fourth most common cancer among women in the United States, with the highest annual incidence rate per 100,000 identified in our local states of New Hampshire and Vermont [31]. Although the majority of diagnosed endometrial cancers are low-grade endometrioid types with early-stage presentation, there is considerable heterogeneity in the histopathologic features, molecular profiles, and prognostic implications associated with these endometrial carcinomas [32, 33]. The four most common subtypes of endometrial carcinoma based on histologic classification are: endometrioid (FIGO grades 1-3), serous, clear cell, and carcinosarcoma.

The FIGO has regularly revised its classification system for endometrial cancer, aiming to better predict disease recurrence and progression, and has recently updated the staging guidelines according to histologic grading [3]. The primary focus of this study is to differentiate between low- and high-grade endometrial cancer which is congruent with these updated guidelines. Accurate identification and classification of endometrial cancer histologic patterns are of paramount importance for patient prognosis and to determine appropriate treatment options. Nevertheless, manual microscopic evaluation can be labor-intensive, and there is the possibility of morphologic overlap between tumor types, leading to some variability among pathologists' interpretations, which may result in suboptimal treatment plans impacting patients' care [34].

As a step towards a solution to these challenges, we have introduced EndoNet, a transformer-based deep learning model developed to classify endometrial cancers into low- or high-grade categories based on their WSIs, thereby assisting in determining prognosis, staging, and treatment [3, 27, 28]. Previous studies have demonstrated the potential of deep learning models to discern the morphologic characteristics of various cancers from histologic images, including endometrial cancer [14-18]. However, these models usually require the manual annotations by pathologists of slides for training and get the inference. In this work, we leveraged the EndoNet model, a method inspired by masked pre-training for histology images using a transformer or MaskHIT, to provide an effective pipeline to analyze WSIs pertinent to endometrial cancer. Although, vision transformers are more computationally expensive than CNNs for the histologic classification in WSIs. But this method eliminates the need for manual annotation of every slide, thereby simplifying the process of getting the second opinion for pathologists in grading endometrial slides without the annotation.

EndoNet employs MaskHIT strategies to capture local and global features and structures within the WSIs. Such a multifactorial analysis is crucial, given that these features collectively contribute to the grading and classification of the disease, which, in turn, guide clinical management. This attribute enables improved visual recognition and aids in distinguishing between low and high-grade cases of endometrial cancer. This approach also highlights the areas within the image that most influence the grading classification, enabling pathologists to see which parts of the image were important in reaching a classification and fostering model utilization and interpretation. The diverse and verifiable foci of EndoNet aim to enhance the precision of histopathological diagnosis and assist pathologists in clinical decision-making.

In this study, we assessed the utility of Fully Supervised CNN approach and EndoNet on 110 patients from our internal testing dataset, as well as on 100 patients from TCGA, which served as an external testing set. Our patch-based CNN model, which required extensive manual annotation by pathologists for training, performed well with good AUC scores on both the internal test set and the external TCGA cohort. On the other hand, the EndoNet model pre-trained on annotated endometrial patches demonstrated superior performance to the model pre-trained on ImageNet on internal test set, except for specificity. This model improved sensitivity from 0.72 to 0.88, although specificity slightly decreased from 0.91 to 0.90. Sensitivity is a particularly critical metric for disease identification as it measures the model's ability to correctly identify those with the disease. These results reflect the model's ability to accurately distinguish between high-grade and low-grade cases based on the analyzed WSIs. Furthermore, EndoNet trained on internal dataset achieved a good performance on the external TCGA cohort, which provided further insight into its performance in a broader and more diverse dataset. These results show a slight decline in performance on the TCGA external dataset compared to the internal test set evaluation. This decline can be attributed to the TCGA dataset's inherent heterogeneity and variability. We observed a pronounced variability in the H&E staining within the TCGA dataset. This is because the slides in the TCGA dataset originate from numerous institutions and are consequently prepared and scanned using diverse protocols and scanners. Despite the inherent heterogeneity and variability in the external dataset, EndoNet maintained its robust performance. Furthermore, incorporating more advanced normalization techniques, such as generative adversarial networks in the pipeline, could potentially address the data variability and enhance performance.

We further investigated the diagnostic performance of EndoNet across various histologic types within both internal and external testing cohorts. For Endometrioid FIGO grade 1 tumors, the model achieved a high accuracy in the internal and external cohorts. EndoNet's moderate diagnostic performance for Endometrioid FIGO grade 2 tumors in both cohorts suggests specific difficulties in correctly identifying these intermediate-grade tumors. This decrease in classification accuracy suggests challenges in distinguishing between the histopathological complexities inherent to the intermediate class relative to the more distinctly characterized grade 1 and grade 3 tumors. This reflects the broader issue within manual pathology grading, where variability in the visual features used by pathologists for grading can lead to discrepancies. This performance is reminiscent of the challenges encountered in manual pathology grading, where the diverse histologic profiles of grade 2 tumors can complicate the grading process. The heterogeneity in the histopathologic presentation of this tumor highlights the need for algorithmic enhancements in future work to refine the specificity of EndoNet, particularly in the crucial differentiation between intermediate and low-grade/high-grade Endometrioid tumors, which is important for accurate clinical prognostication and therapeutic decision-making. Additionally, we aim to refine EndoNet by investigating the transition from Endometrioid FIGO grade 1 to grade 2, which is traditionally based on the percentage of solid growth and the presence of significant nuclear atypia. Understanding and integrating these visual grading features can potentially improve the EndoNet's grading accuracy of Endometrioid FIGO grade 2 tumors and hence improving the diagnostic and prognostic outcomes.

For Serous Carcinoma, EndoNet's accuracy, while good accuracy in both test sets, might still benefit from further training, particularly considering that serous carcinomas are high-grade by definition and possess gland-like structures and have overlapping architectural features with the grade 1-2 endometrioid tumor. However, serous carcinomas typically show more severe cell abnormalities. The potential for confusion with low-grade endometrioid cancers, which may share some features with serous carcinomas, could be an area for future improvement. Of note, EndoNet showed high accuracy in diagnosing carcinosarcoma, a complex and relatively rare malignancy characterized by both carcinomatous and sarcomatous components. Although the prevalence of this type in the internal dataset is very low, the algorithm's ability to correctly predict 7 out of 8 patients in the internal test set is notable. However, it is crucial to acknowledge the absence of carcinosarcoma cases in the TCGA dataset for assessment. These evaluation results indicate that while EndoNet holds promise as a diagnostic tool, it may still benefit from further improvement in more nuanced histologic subtypes, especially for grade 2 endometrioid and serous carcinomas. Overall, this analysis highlights EndoNet's potential as a reliable tool for classifying histologic subtypes of endometrial cancer.

The qualitative investigation of our model visualizations, such as those in Figure 3, showed that the model exhibits an increased focus or 'attention' towards regions that heavily overlap with endometrial cancer tissues independently segmented by expert pathologists. Although, the scope of this analysis is rather limited as this was performed on very limited cases rather than the whole internal and external test sets. The primary aim of this analysis was to determine if the attention of EndoNet is accurately focus on the tumoral regions, which was confirmed by the pathologists. In the future, we plan to expand this analysis to whole dataset to develop it as a useful tool which can enhance its interpretability and trustworthiness by providing visual and tangible evidence for its classification output, offering valuable insights to pathologists.

Our research has set a foundation for future explorations into implementing computational pathology tools for endometrial cancer diagnostics. However, several limitations of the EndoNet study must be acknowledged. The study's emphasis on histologic classifications does not encompass the molecular aspects or the less prevalent histologic types like clear cell carcinoma, dedifferentiated/undifferentiated carcinomas, and tumors with mixed histologic variants, potentially limiting a comprehensive classification of endometrial cancer [3]. The focus on morphological features extracted from WSIs without integrating clinical, genomic, and molecular features might restrict the potential to enhance diagnostic accuracy further. The current study design does not include prognosis prediction, an important component in cancer treatment planning. Acknowledging the importance of molecular classification in the comprehensive staging of endometrial cancer, we plan to extend our methodology in future work to incorporate both molecular and histological data and develop a multi-modal approach for comprehensive risk stratification and prognosis. While the shift towards molecular-based diagnosis of endometrial cancer offers advantages in subtype differentiation and grading with reduced interobserver variability, our study underscores the utility of deep learning in histological image analysis, particularly when molecular testing is not accessible. This is crucial in low-resource settings, where molecular diagnostics may be limited. Our current approach, focusing on the latest FIGO staging 2023, serves as a proof of concept that histology image analysis can effectively classify low- and high-grade endometrial cancers.

Another limitation of our study is that our model overlooks other significant prognostic features like lymphovascular space invasion. In addition, we were not able to specifically evaluate the algorithm's interpretation of squamous metaplasia in the classification of endometrioid tumors. The squamous metaplasia is a frequent occurrence in endometrioid cancers, can have a solid growth pattern but is not included as contributing to the solid component in the grading of endometrioid cancers. Future studies could potentially explore whether including squamous metaplasia when classifying these cancers with EndoNet could change how the tumors are graded, especially if there is a higher proportion of squamous differentiation. These limitations delineate areas for improvement and expansion, laying a roadmap for future investigations in this domain.

We envision several directions for future work and further investigation. To enhance the applicability of our model, it is crucial to assess its performance across various histologic subtypes of endometrial cancer, which can provide a broader perspective on EndoNet's capabilities and help refine the model for better performance across a more comprehensive array of cases. The exclusion of Clear Cell Carcinoma (CCC) from our study was due to the extreme rarity of this subtype, which limited the number of cases available in our dataset for model development and evaluation. We plan to include additional high-grade classes such as CCC, mesonephric-like carcinomas, and undifferentiated carcinomas, by seeking out more datasets and exploring partnerships with other institutions to increase our case numbers for these high-grade classes. We also plan to focus on differentiating the classification of the high-grade tumor subtypes. We plan to incorporate other relevant genetic information, immunohistochemistry staining results (e.g., *p53*, *p16*, *ER*, *PR*), and relevant demographic factors into our model's training pipeline. The goal is to equip our model to account for the complex interplay of genetic and clinical information in endometrial cancer prognostics, thereby enhancing its predictive accuracy.

Furthermore, we aim to introduce EndoNet as part of a clinical decision-support system, empowering pathologists with a new tool to improve diagnostic accuracy, expedite decision-making, and enhance patient care for endometrial cancer. To achieve this goal, we intend to conduct a prospective clinical trial to quantify the impact of EndoNet on the performance of pathologists and patient health outcomes, employing suitable clinical metrics to measure its potential benefits in real-world settings, including its effects on diagnostic efficiency, accuracy, and patient treatment outcomes. The deployment of EndoNet in clinical settings and the measurement of the reduction in interobserver variability among pathologists represent definitive methods to assess the clinical impact of our approach. However, given the resources required for such an evaluation, this aspect is currently beyond the scope of our study. We plan to conduct a follow-up prospective clinical trial to compare the model's performance with that of pathologists and to evaluate its clinical impact, specifically regarding the enhancement of pathologists' performance and the reduction of interobserver variability, as part of our future work. We have strived to demonstrate the potential benefits of this approach over manual methods in terms of accuracy in grading endometrial cancer, pending a future prospective clinical trial. These approaches can be especially beneficial in settings where access to subspecialty pathologists is limited. By providing promising retrospective internal and external evaluations, this work lays the groundwork for future prospective evaluations of deploying our method in clinical settings. In the future, we plan to collaborate further with gynecologic pathologists and other institutions to refine and evaluate our model further. We envision that the model could provide a robust second opinion to pathologists and enhance the efficiency of their diagnostic performance.

**Table 1.** Distribution of classes in the internal and TCGA datasets.

| Histologic Subtype | No. of slides in Internal Dataset | No. of slides in TCGA dataset |
|---|---|---|
| Endometroid FIGO grade 1 | 402 | 48 |
| Endometroid FIGO grade 2 | 252 | 22 |
| Endometroid FIGO grade 3 | 109 | 24 |
| Serous Carcinoma | 119 | 6 |
| Carcinosarcoma | 47 | NA |
| *Total* | *929* | *100* |

**Table 2.** Distribution of number of WSIs and patients across different partitions for model development and evaluation.

| Histologic Grading | Internal Training Dataset | Internal Validation Dataset | Internal Testing Dataset | External Testing Dataset (TCGA) | Total |
|---|---|---|---|---|---|
| Low-grade cases | 456 slides from 269 patients | 65 from 42 patients | 132 from 78 patients | 70 from 70 patients | 724 from 459 patients |
| High-grade cases | 192 slides from 85 patients | 23 from 9 patients | 56 from 32 patients | 30 from 30 patients | 305 from 156 patients |
| *Total* | *648 slides from 354 patients* | *93 from 53 patients* | *188 from 110 patients* | *100 from 100 patients* | *1029 from 615 patients* |

**Table 3.** Evaluation of the performance of proposed EndoNet on the internal and external test set. The 95% CI of each metric is included in parentheses.

| Methods | Testing set | F1 Score | AUC Score | Sensitivity | Specificity |
|---|---|---|---|---|---|
| Fully Supervised CNN (ResNet18) | Internal test set | 0.78 (0.70–0.86) | 0.82 (0.72–0.90) | 0.90 (0.79–1.00) | 0.67 (0.56–0.77) |
| | TCGA test set | 0.86 (0.80–0.92) | 0.79 (0.68–0.89) | 0.57 (0.38–0.74) | 0.90 (0.83–0.97) |
| EndoNet Pre-trained on ImageNet | Internal test set | 0.90 (0.85–0.95) | 0.87 (0.78–0.95) | 0.72 (0.56–0.87) | 0.91 (0.84–0.97) |
| | TCGA test set | 0.79 (0.68–0.90) | 0.83 (0.74–0.91) | 0.87 (0.61–1.00) | 0.69 (0.53–0.93) |
| EndoNet Pre-trained on Endometrial Patches | Internal test set | 0.91 (0.86–0.95) | 0.95 (0.89–0.99) | 0.88 (0.76–0.97) | 0.90 (0.83–0.96) |
| | TCGA test set | 0.86 (0.80–0.94) | 0.86 (0.75–0.93) | 0.87 (0.69–0.97) | 0.80 (0.71–0.89) |

**Table 4.** Evaluation performance of proposed EndoNet for classification of the histologic subtypes on the internal and external test sets.

| Diagnosis/ Dataset | Accuracy on the internal test set (No. correct predictions/ No. of patients) | Accuracy on TCGA/external test set (No correct predictions/ No. of patients) |
| --- | --- | --- |
| Endometroid FIGO grade 1 | 0.94 (48/ 51) | 0.85 (41/ 48) |
| Endometroid FIGO grade 2 | 0.81 (22/ 27) | 0.68 (15/ 22) |
| Endometroid FIGO grade 3 | 0.91 (10/ 11) | 0.88 (21/ 24) |
| Serous Carcinoma | 0.85 (11/ 13) | 0.83 (5/ 6) |
| Carcinosarcoma | 0.88 (7/ 8) | NA |

**Figure Legends:**

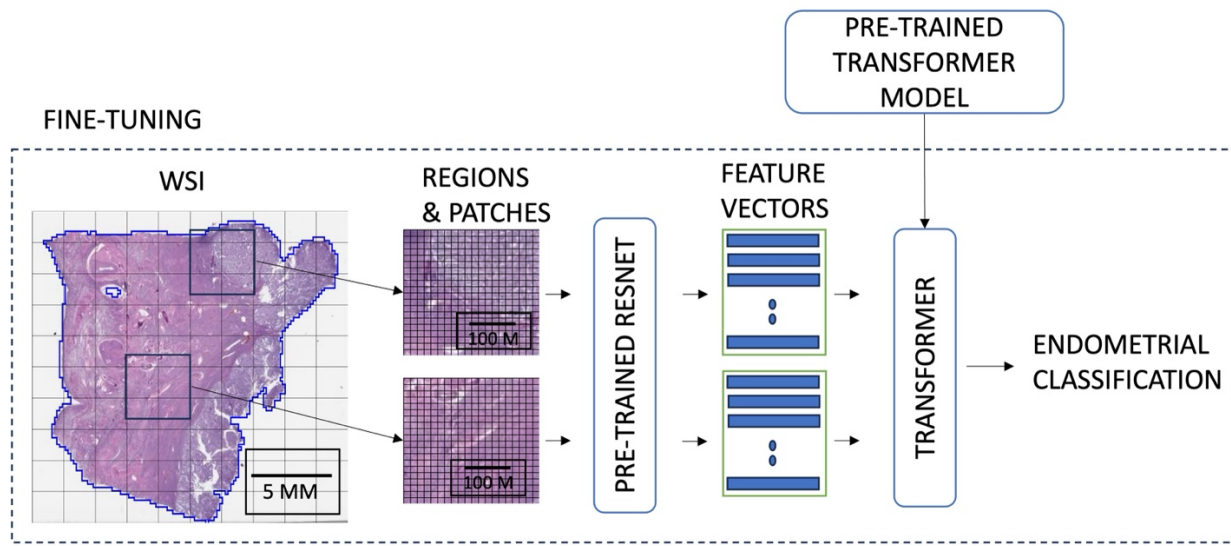

**Figure 1.** The overview of fine-tuning the EndoNet Architecture for classification of Endometrial Cancers on H&E slides.

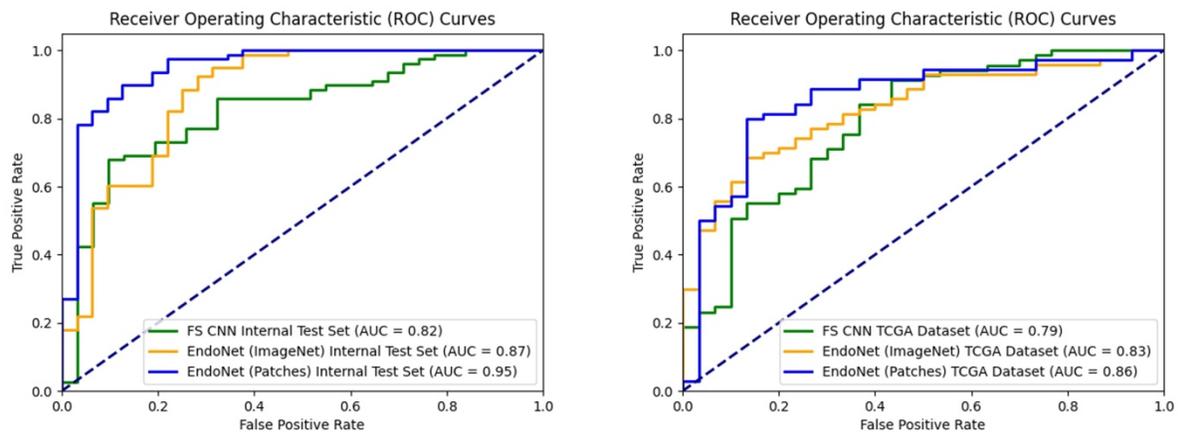

**Figure 2.** ROC curves demonstrate the performance of Fully Supervised CNN approach (FS CNN) and EndoNet on both the internal and the TCGA (external) test sets. The green lines represent FS CNN, the orange lines depict the EndoNet model pre-trained ResNet on the ImageNet dataset, and the blue lines represent the EndoNet model utilizing ResNet pre-trained on endometrial patches.

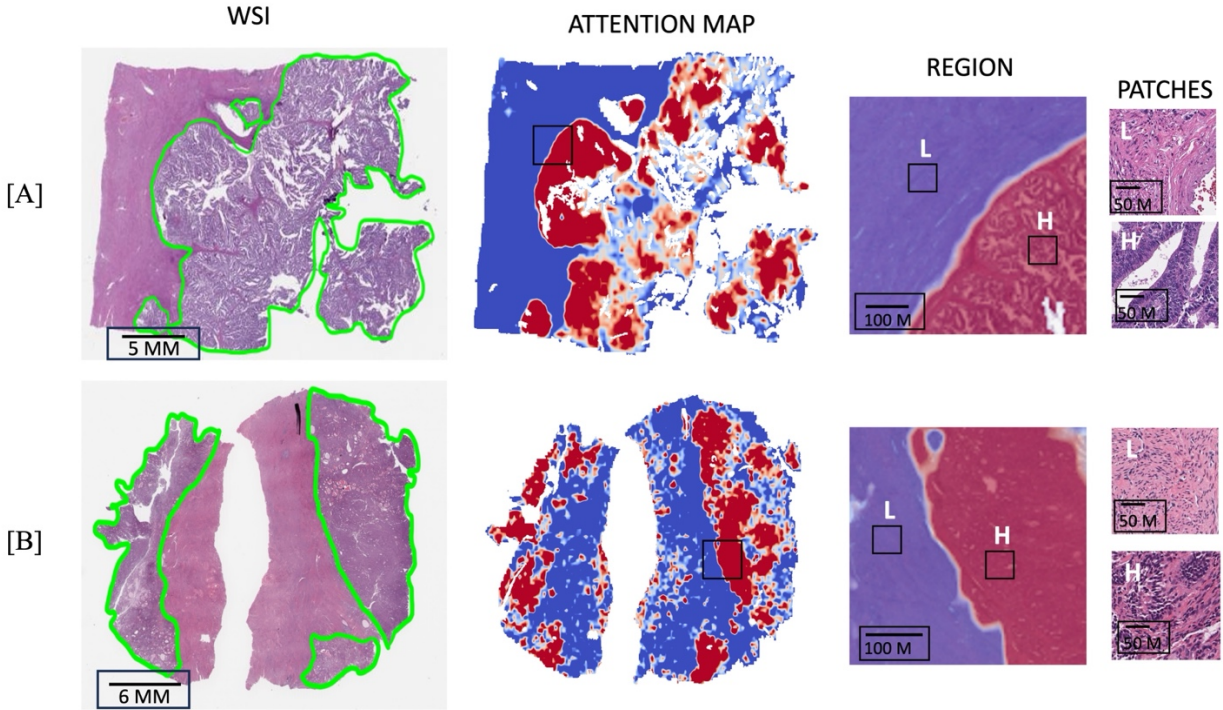

**Figure 3.** Display of attention maps in endometrial cancer. The region size is 4480μm, and the patch size is 224μm. This figure illustrates the model's attention areas in a typical histology slide (A: Internal Test set, B: External TCGA test set), represented in panels 1) to 4). The high-attention areas are marked in red (H), while the areas of lower attention are indicated in blue (L). The first column, labeled "WSI," shows the original whole slide image with manual annotations by the pathologist. The tumor regions are delineated in green.